\documentclass[runningheads]{llncs}
\usepackage[T1]{fontenc}
\usepackage{graphicx}
\usepackage{tabularx}
\usepackage{cite}
\usepackage{amsmath,amssymb,amsfonts}
\usepackage{algorithmic}
\usepackage{graphicx}
\usepackage{textcomp}
\usepackage{xcolor}
\usepackage{academicons}
\usepackage[misc, geometry]{ifsym}
\usepackage{siunitx}
\usepackage{booktabs}
\usepackage{multirow}
\usepackage{svg}
\usepackage{amsmath}
\usepackage{wrapfig}
\usepackage{sidecap}
\newbox{\myorcidaffilbox}
\sbox{\myorcidaffilbox}{\large\includegraphics[height=1.25ex]{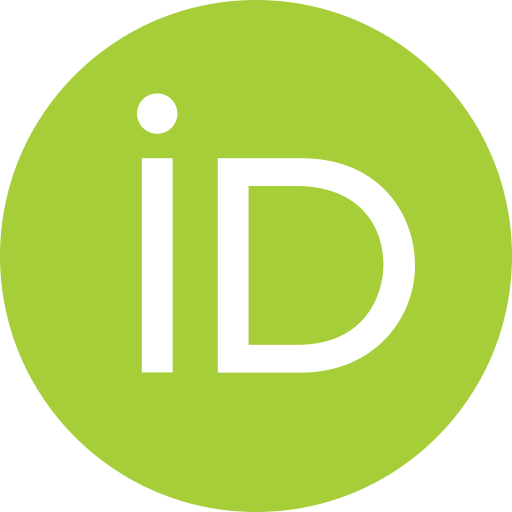}}
\newcommand{\orcidaffil}[1]{%
  \href{https://orcid.org/#1}{\usebox{\myorcidaffilbox}}}

\newcommand{\comment}[1]{\ignorespaces}

\usepackage{hyperref}
\usepackage{color}

\def\BibTeX{{\rm B\kern-.05em{\sc i\kern-.025em b}\kern-.08em
    T\kern-.1667em\lower.7ex\hbox{E}\kern-.125emX}}

\begin{document}

\title{\LARGE \bf SLUGBOT, an \emph{Aplysia}-inspired Robotic Grasper for Studying Control\thanks{This work was supported by NSF DBI2015317 as part of the NSF/CIHR/DFG/FRQ/UKRI-MRC Next Generation Networks for Neuroscience Program and by the NSF Research Fellowship Program under Grant No. DGE1745016. Any opinions, findings, and conclusions or recommendations expressed in this material are those of the authors and do not necessarily reflect the views of the National Science Foundation.}
}

\titlerunning{SLUGBOT, an \emph{Aplysia}-inspired Robotic Grasper}

\author{
    Kevin Dai\inst{1}$^{+}$\textsuperscript{\orcidaffil{0000-0002-5895-0450}} \and
    Ravesh Sukhnandan\inst{2}$^{+}$\textsuperscript{\orcidaffil{0000-0001-5858-961}} \and
    Michael Bennington\inst{1}$^{+}$\textsuperscript{\orcidaffil{0000-0001-7940-6297}} \and
    Karen Whirley\inst{1}\textsuperscript{\orcidaffil{0000-0002-7914-1564}} \and
    Ryan Bao\inst{1}\textsuperscript{\orcidaffil{0000-0003-4548-7687}} \and
    Lu Li\inst{4}\textsuperscript{\orcidaffil{0000-0002-3346-283X}} \and Jeffrey P. Gill\inst{5}\textsuperscript{\orcidaffil{0000-0002-4115-8045}} \and
    Hillel J. Chiel\inst{5,6,7}\textsuperscript{\orcidaffil{0000-0002-1750-8500}} \and
    Victoria A. Webster-Wood\inst{1,2,3}\textsuperscript{\orcidaffil{0000-0001-6638-2687}\Letter} 
}

\authorrunning{K. Dai et al.}

\institute{$^1$Dept. of Mechanical Engineering,
$^2$Dept. of Biomedical Engineering\\
$^3$McGowan Institute for Regenerative Medicine, $^4$Robotics Institute\\Carnegie Mellon University, Pittsburgh, PA, United States\\ \hspace{1pt} \\
$^5$Department of Biology, $^6$Department of Neurosciences\\
$^7$Department of Biomedical Engineering\\ Case Western Reserve University, Cleveland, OH, United States\\ \hspace{1pt} \\
$^{+}$These authors contributed equally to the work.\\
\email{vwebster@andrew.cmu.edu}
}

\maketitle

\vspace{-20pt}
\begin{abstract}

Living systems can use a single periphery to perform a variety of tasks and adapt to a dynamic environment. This multifunctionality is achieved through the use of neural circuitry that adaptively controls the reconfigurable musculature. Current robotic systems struggle to flexibly adapt to unstructured environments. Through mimicry of the neuromechanical coupling seen in living organisms, robotic systems could potentially achieve greater autonomy.  The tractable neuromechanics of the sea slug \emph{Aplysia californica's} feeding apparatus, or buccal mass, make it an ideal candidate for applying neuromechanical principles to the control of a soft robot. In this work, a robotic grasper was designed to mimic specific morphology of the \textit{Aplysia} feeding apparatus. These include the use of soft actuators akin to biological muscle, a deformable grasping surface, and a similar muscular architecture. A previously developed Boolean neural controller was then adapted for the control of this soft robotic system. The robot was capable of qualitatively replicating swallowing behavior by cyclically ingesting a plastic tube.  The robot's normalized translational and rotational kinematics of the odontophore followed profiles observed \textit{in vivo} despite morphological differences. This brings \emph{Aplysia}-inspired control \emph{in roboto} one step closer to multifunctional neural control schema \emph{in vivo} and \emph{in silico}. Future additions may improve SLUGBOT's viability as a neuromechanical research platform.

\vspace{-10pt}

\end{abstract}


\vspace{-5 pt}
\section{INTRODUCTION}
\vspace{-15 pt}
Robots have typically struggled to replicate the innate behavioral flexibility of animals to act in unstructured environments \cite{royakkers_literature_2015, valero-cuevas_neuromechanical_2017}. Animals can fluidly adapt to their changing environment by exploiting the close coupling of their neural and muscular systems \cite{lyttle_robustness_2017, nishikawa_neuromechanics_2007}.  This behavioral flexibility is driven in part by the multifunctionality of limbs and appendages, which can achieve different tasks by enacting new motor control strategies or leveraging changing mechanical advantages \cite{lyttle_robustness_2017,nishikawa_neuromechanics_2007}.  One bioinspired method specifically integrates knowledge of an organism’s neuromechanics to improve robotic performance in tasks such as locomotion and grasping~\cite{mangan2005biologically, nishikawa_neuromechanics_2007}. Here, knowledge of the model animal is used in the design of the robot’s morphology and control. These bio-inspired robots may also serve as a platform for testing biological hypotheses and validating neuromechanical simulation since such robots must solve real physical problems~\cite{nishikawa_neuromechanics_2007,pfeifer_how_2006}.  

\begin{figure}[b]
    \centering
    \includegraphics[width=\textwidth]{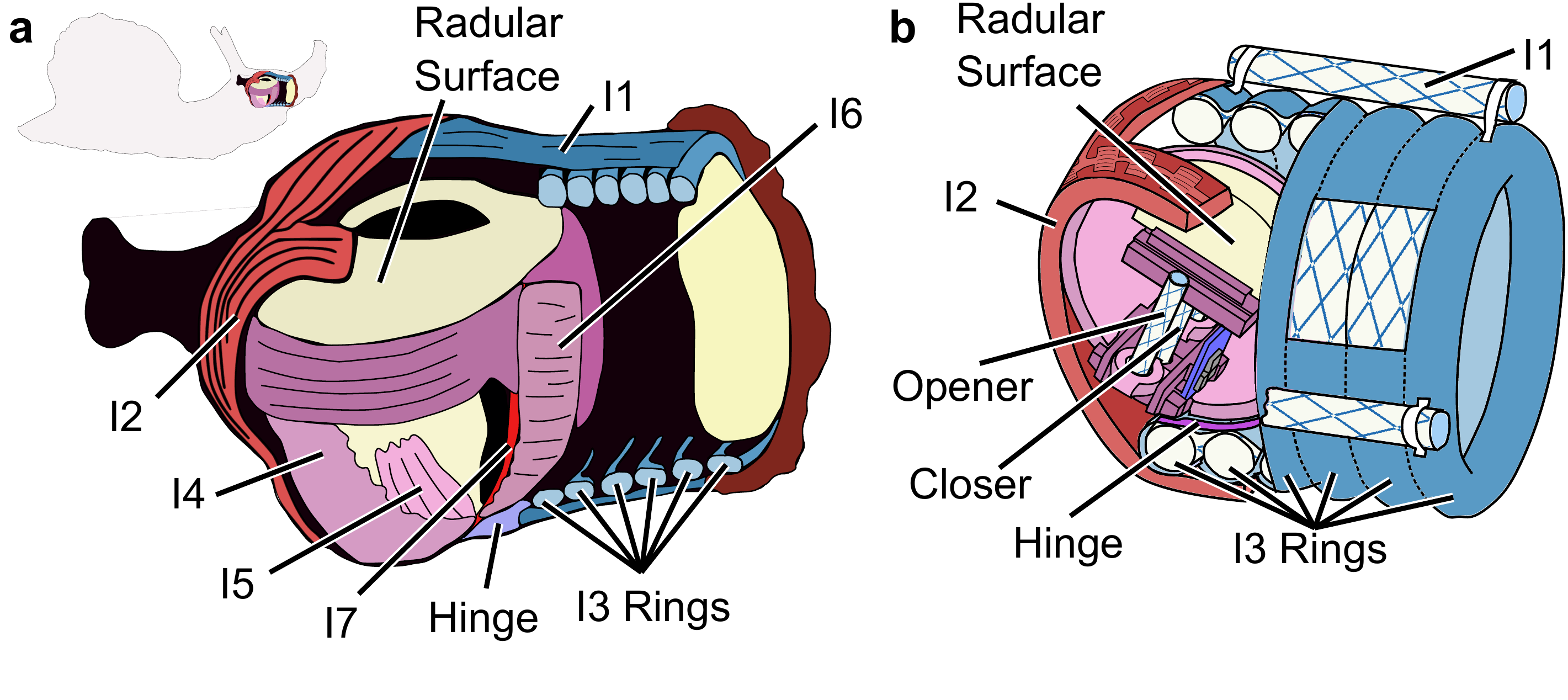}
    \vspace{-30pt}
    \setlength{\belowcaptionskip}{-15pt}
    \caption{Cutaway views of a) \emph{Aplysia} buccal mass anatomy located in the animal (modified with permission \cite{Drushel1998KinematicCalifornica}) and b) SLUGBOT, showing key muscle groups and features.}
    \label{fig:overview}
\end{figure}

Determining the right level of biological detail to capture in bioinspired and biomimetic robotics for neuromechanical research remains an ongoing challenge~\cite{ritzmann_convergent_2004}. In addition to the difficulties in matching the mechanical properties of muscles and tissues, replicating the complex neural circuits that control the periphery for robotic control is often impossible as such circuits are not known in their entirety \cite{ritzmann_convergent_2004}. The neural circuitry governing the feeding behavior of the marine mollusk \textit{Aplysia californica} has been extensively studied \cite{Webster-wood2020ControlCalifornica}, which makes it an ideal model organism for investigating the level of biomimicry needed to create a robotic platform for neuromechanical research.  In this work, we present the real-time control of a soft-robotic Slug-Like Uniaxial Grasper roBOT (SLUGBOT) inspired by the morphology and control of \textit{Aplysia’s} feeding apparatus (buccal mass, Figure \ref{fig:overview}a).  The grasper is controlled by a modified version of Webster-Wood et al.’s hybrid Boolean model of \textit{Aplysia} neuromechanics \cite{Webster-wood2020ControlCalifornica}, and can qualitatively replicate \textit{Aplysia} swallowing behavior.  This robot may serve as a platform for testing future hypotheses related to \textit{Aplysia’s} neuromechanics, as well as for elucidating techniques for generating multifunctional grasping behavior.  

\vspace{-5 pt}
\subsection{Feeding behavior in \textit{Aplysia} and prior work}
\vspace{-5 pt}
\textit{Aplysia} moves food through its buccal mass primarily by means of a grasper composed of a layer of flexible cartilage covered with very fine teeth (radula) that covers a muscular structure (odontophore) \cite{kehl_soft-surface_2019}.  The grasper closes to grasp food and opens to release it, and can either protract (pivot and translate towards the animal's jaws) or retract (pivot and translate towards the esophagus and crop)\cite{gill_rapid_2020}.  \textit{Aplysia} is capable of generating multifunctional behavior by varying the timing of the activation of muscles responsible for protraction, retraction, and opening and closing of the grasper \cite{morton_vivo_1993, neustadter_kinematics_2007}. In the ingestive behaviors of biting and swallowing, the grasper is closed during retraction to bring food into the esophagus and crop~\cite{gill_rapid_2020}.  

The previous state-of-the-art in \textit{Aplysia} inspired robot was developed by Mangan et al., who created a pneumatically actuated soft-robotic gripper inspired by \textit{Aplysia’s} feeding apparatus morphology \cite{mangan2005biologically}.  The device consisted of McKibben actuators that activated to peristaltically protract or retract a rubber ellipsoid.  The rubber ellipsoid grasped objects by pressurizing McKibben actuators that served as inflatable “lips”. Although this robot included the I3 retractor muscle, it did not include several key muscles involved in feeding behavior in the animal, namely  the I2 muscle, which is the main driver of protraction \cite{Yu1999BiomechanicalAplysia}, the I1 muscle, and separate opener and closer muscles for the radular surface \cite{kehl_soft-surface_2019}.

The peristaltic sequence used to protract and retract Mangan et al.'s grasper did not incorporate a neural controller based on the established neural circuitry of \textit{Aplysia}.  It also did not include real time sensory feedback to trigger multifunctional behavior based on changing stimuli. Our robot addresses this gap by implementing real time control of the robot's actuators based on the Boolean model of \textit{Aplysia}'s neuromechanics by Webster-Wood et al. \cite{Webster-wood2020ControlCalifornica}. The Boolean model uses simplified biomechanics and an on/off representation for neuronal activity \cite{Webster-wood2020ControlCalifornica}, which allows faster than real time simulation of the model.  These properties make it suitable for real-time control of the robotic grasper.
 
 \vspace{-5 pt}
\section{Methods}
\vspace{-5 pt}

SLUGBOT is a robotic representation of the buccal mass of \textit{Aplysia}. We represent the odontophore and a portion of surrounding musculature including the I1, I2 and I3 muscles (Figure \ref{fig:overview}b).  The hinge is represented by an elastic band.

\begin{figure*}[ht!]
    \centering
    \includegraphics[width=\textwidth]{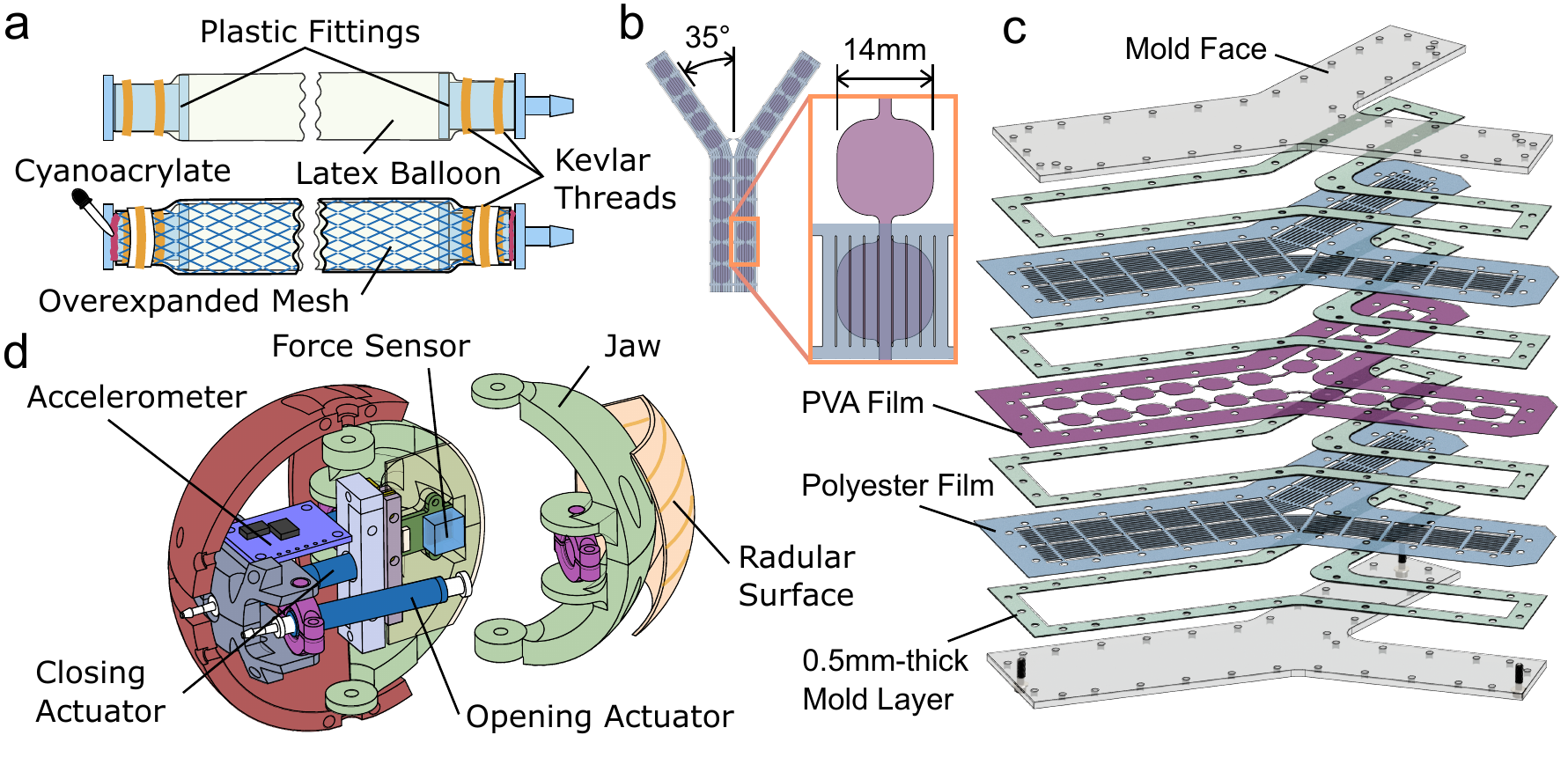}
    \vspace{-25pt}
    \setlength{\belowcaptionskip}{-15pt}
    \caption{Fabrication of SLUGBOT actuators and grasper. a) Construction of McKibben actuators with internal layer (upper) and external layer (lower), b) Y-shaped I2 muscle geometry with PVA film dimensions and PET reinforcing film, c) I2 FPAM mold and material layup during casting process, and d) cutaway view with partial exploded view of odontophore grasper geometry.}
    \label{fig:manufacturing}
\end{figure*}

\subsection{Fabricating McKibben artificial muscles}
The I1 muscles (\SI{125}{\milli\metre}-long), and the odontophore opener and closer muscles (\SI{45}{\milli\metre}-long, see Section~\ref{sec:odontophore}) were represented by linear McKibben actuators, while the I3 muscle was represented by a series of unsegmented, toroidal McKibben actuators (\SI{295}{\milli\metre} circumference). Both types of McKibbens share a similar fabrication scheme (Figure~\ref{fig:manufacturing}a). The inner bladder of the McKibben consisted of a latex balloon attached to plastic fittings. The linear McKibbens required one barbed \SI{3.18}{\milli\metre} to \SI{1.59}{\milli\metre} reducer fitting and one \SI{3.18}{\milli\metre} plug fitting, whereas the toroidal McKibbens required two barbed reducers. The latex balloon was held onto the fittings with two Kevlar thread slipknots to create an airtight seal. A soapy water test was performed on the inner bladder to ensure it was sealed before adding the outer mesh. A \SI{9.53}{\milli\metre} overexpanded braided sleeving (Flexo\textsuperscript{\tiny\textregistered} PET) was heated at both ends to prevent unravelling of the mesh, and then held in place over the latex bladder using Kevlar slipknots and \SI{6}{\milli\metre}-long rings of latex balloons, which were placed between the mesh and the Kevlar thread to increase friction and prevent slipping. To complete the linear McKibben actuators, cyanoacrylate (CA) adhesive (Loctite Gel) was used to adhere the ends of the braided sleeve to the fittings and the latex bladder. To form the circular shape of the toroidal muscles, the two barbed reducers were also connected by \SI{12}{\milli\metre} lengths of tubing to a \SI{1.59}{\milli\metre} Y-fitting.

\vspace{-10pt}
\subsection{Fabricating flat artificial muscles}
\vspace{-5 pt}
The I2 muscle was represented by a \SI{2}{\milli\metre}-thick, Y-shaped assembly of flat pneumatic artificial muscles (FPAMs) containing two series of \SI{20}{\milli\metre}-wide FPAM cells (Figure~\ref{fig:manufacturing}b)~\cite{wirekoh2017design,wirekoh2019sensorized,park2020organosynthetic}. The FPAMs were structured with \SI{25.4}{\micro\metre}-thick polyester (PET) reinforcing films on each side of a polyvinyl alcohol masking film (PVA, Sulky Ultra Solvy), separated by \SI{0.5}{\milli\metre}-thick layers of elastomer (Smooth-On Dragonskin 10 Slow). The films were cut using a Silhouette Portrait 2. Creases in the PVA film were smoothed out prior to cutting by annealing the film under tension with a heat gun, taking care to avoid warping the PVA film. The PET reinforcing film was wiped after cutting with isopropyl alcohol and DOWSIL 1200 OS primer, which was allowed to cure for 60-90 minutes prior to elastomer casting. For forming the elastomer, two laser-cut pieces of acrylic formed the top and bottom faces of a mold along with four \SI{0.5}{\milli\metre}-thick Y-shaped outlines 3D-printed with polylactic acid (PLA) to form a Y-shaped cavity when placed between the mold's acrylic faces. The FPAMs were fabricated by sequentially (1)~assembling the mold in \SI{0.5}{\milli\metre} layer increments on the bottom acrylic plate, (2)~casting liquid elastomer into the mold, (3)~smoothing elastomer using a wooden mixing stick, and (4)~adding reinforcing or masking films onto the elastomer as necessary, and (5)~repeating these steps until the casting reached a total thickness of \SI{2}{\milli\metre} (Figure~\ref{fig:manufacturing}c). M3 fasteners indexed the assembly of the mold and films during casting, taking care to avoid entrapping air bubbles in the elastomer when tightening the top acrylic face. After the elastomer was cured and excess material trimmed, de-ionized water was flushed through the FPAM cavities with a syringe to remove the PVA film. Once the I2 muscle dried, \SI{3.18}{\milli\metre} barbed fittings were adhered into the cavity ends using SmoothOn Sil-Poxy.

\vspace{-10 pt}
\subsection{Odontophore design and fabrication}
\label{sec:odontophore}
\vspace{-5 pt}

\begin{wrapfigure}[12]{r}{0.4\textwidth}
    \vspace{-32pt}
    \centering
    \includegraphics[width=.4\textwidth]{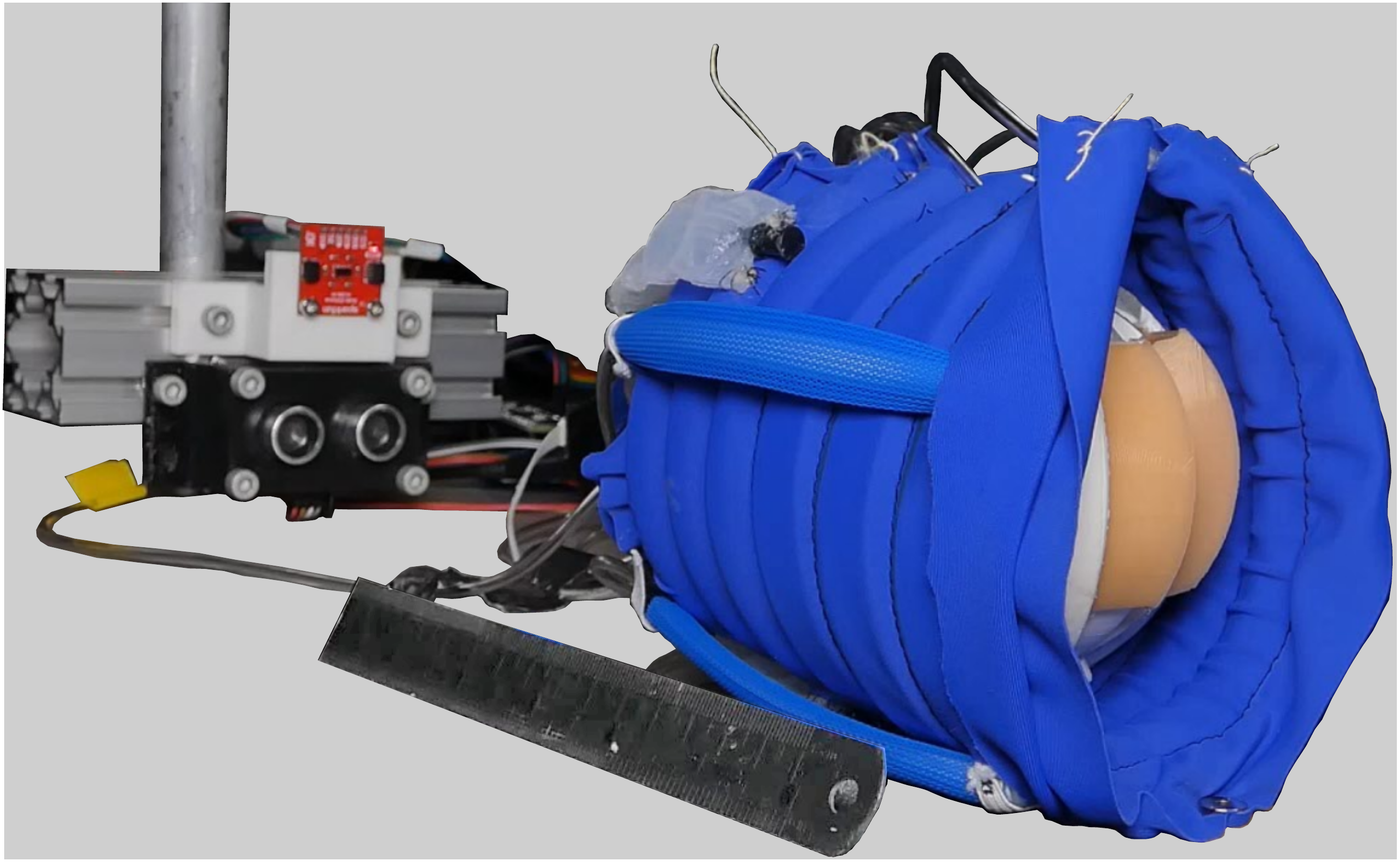}
    \vspace{-20pt}
    \caption{SLUGBOT experimental setup showing the odontophore in the forward position and the laser time-of-flight sensor positioned at the rear of the robot.}
    \label{fig:slugrobotexperiment}
\end{wrapfigure}

The odontophore grasper was designed as a functional analogy to the \textit{Aplysia} odontophore instead of as a direct anatomical model. The key functions include a soft gripping surface that can be opened bilaterally using one set of muscles and closed with variable gripping pressure using antagonist muscles (Figure~\ref{fig:manufacturing}d). The grasper consisted of a rigid PLA outer shell fabricated by 3D printing, two 3D printed radular halves, and a cast rubber surface. The radular halves pivoted freely about two M3 bolts at the shell's poles. The grasper was actuated by three linear McKibben actuators forming one antagonistic pair. Two of the muscles shared an air inlet and actuated together to open the grasper. These muscles were attached to the internal edge of the radular halves and to the back of the shell. The third McKibben was placed in the middle of the grasper and attached to the internal edge of the soft surface with a 3D printed clamp. When activated, this muscle pulled the surface further into the grasper and pulled the radular halves together to generate closing pressure. 

The soft gripping surface was fabricated by casting rubber (SmoothOn Vytaflex 30) into a 3D printed mold lined with circumferential Kevlar threads, creating an inextensible surface in the closer muscle's pulling direction. The surface was attached to the radular halves using CA glue applied only at the edge distal to the opening, which allowed the surface to lift away from the radula when the  closer muscle was inactive but remained taut with an activated closer muscle.

The grasper was fitted with two sensors: 1) a 9-degree of freedom inertial measurement unit (BNO055) to sense absolute orientation of the grasper, and 2) a Hall-effect-based soft magnetic force sensor~\cite{harber2020tunable,DaiTactile2021} to measure the grasper's closing pressure. A laser time-of-flight sensor (VL53L5CX, SparkFun\textsuperscript{\tiny\textregistered}) was mounted externally and pointed at the odontophore outer shell to obtain positional feedback. Retroreflective tape was applied to the external surface of the shell where the laser was expected to coincide. The remainder of the shell was coated with \SI{76.2}{\micro\metre}-thick Ultra-High Molecular Weight tape to reduce friction.

\vspace{-5 pt}
\subsection{Robot assembly}
\vspace{-5 pt}

The actuators representing the I1, I2, and I3 muscles were assembled using a nylon spandex sheath. Two sheets of spandex fabric were sewn together with \SI{300}{\milli\metre}-long, parallel seams separated by \SI{20}{\milli\metre}, which formed six channels for housing the I3 toroidal actuators. The linear I1 actuators were attached on the outside of the sheath with \SI{6.35}{\milli\metre}-wide elastic straps over fittings at each end of each actuator. The straps were positioned to avoid constricting the actuator when inflated. The I2 actuator was attached at each corner to the posterior portion of the spandex sheath with thread. Sil-Poxy was applied to the thread contacting the I2 actuator to avoid tear-out of the silicone actuator. A \SI{6.35}{\milli\metre}-wide elastic band representing the hinge was attached at one end to the posterior I3 ring using a hook-and-loop fastener, and attached to the odontophore at the other end using a sewn loop of kevlar thread that wrapped around a pole bolt.

\begin{wrapfigure}[15]{r}{0.54\textwidth}
    \vspace{-50pt}
    \centering
    \includegraphics[width=.54\textwidth]{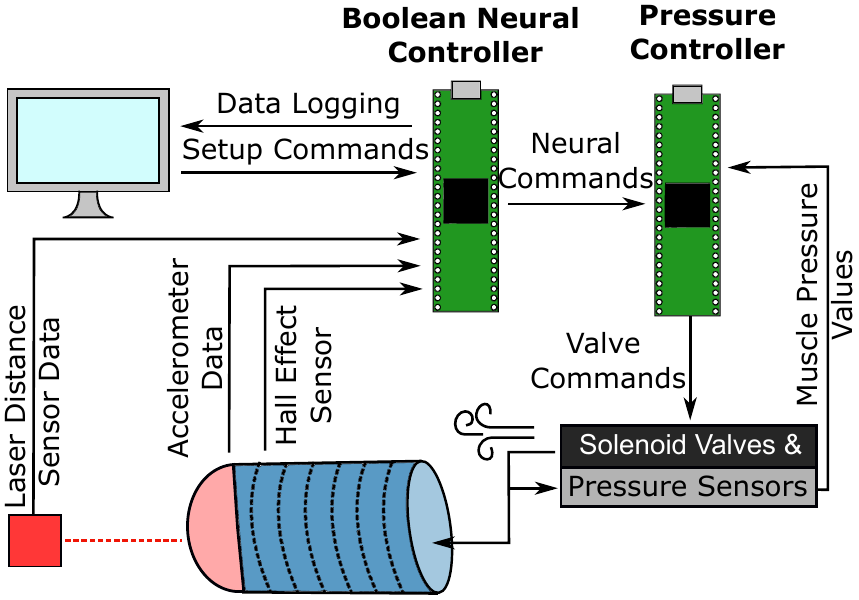}
    \setlength{\abovecaptionskip}{-15pt}
    \caption{SLUGBOT control schema. Behavioral stimuli and sensor feedback are input to the neural controller, which outputs neural signals to the pressure controller. The pressure controller integrates neural signals into muscle activation pressures.}
    \label{fig:schematic}
\end{wrapfigure}

\subsection{Robot control}

To qualitatively compare the robot's movements to those of the published hybrid Boolean model \cite{Webster-wood2020ControlCalifornica}, we prepared the robot to accept simulated Boolean neural commands (Figures~\ref{fig:slugrobotexperiment} and \ref{fig:schematic}). The neural commands were generated by one microcontroller and sent to a second microcontroller that integrated the neural command signals into muscle activations and controlled the pressure within each actuator. Prior to each experiment, the odontophore was placed in the posterior of the robot touching the I2 muscle, with radula facing upwards. Behavioral movement was initialized by pushbuttons corresponding to seaweed-inspired stimuli.

\subsubsection{Boolean Neural Controller}

A modified version of the Boolean neural network presented by Webster-Wood et al. \cite{Webster-wood2020ControlCalifornica} was implemented on a microcontroller (Teensy 3.6) to determine the muscle activation patterns in real time. This model represents bursting neural activity as discrete Boolean variables and the neural interactions are captured in logical calculations.

\begin{table}[htbp]
\caption{Modifications to Hybrid Boolean Model for SLUGBOT}
\begin{center}
\begin{tabularx}{\textwidth}{
>{\centering\arraybackslash}m{\dimexpr.16\linewidth-2\tabcolsep-1.3333\arrayrulewidth}
>{\centering\arraybackslash}m{\dimexpr.35\linewidth-2\tabcolsep-1.3333\arrayrulewidth}
m{\dimexpr.49\linewidth-2\tabcolsep-1.3333\arrayrulewidth}
}
\toprule
{\textbf{Motor Unit}} & {\textbf{Activated Actuator Group}} & \multicolumn{1}{>{\centering\arraybackslash}m{\dimexpr.5\linewidth-2\tabcolsep-1.3333\arrayrulewidth}}{\textbf{Activation Logic}}\\
\midrule
\midrule
RU1 & Anterior/medial I3 & Same as B3/B6/B9 motor pool in \cite{Webster-wood2020ControlCalifornica}\\[3pt]
\multirow{2}{*}{RU2/RU3} & \multirow{2}{*}{Medial/posterior I3} & Fixed time delays after Retractor Unit 1\\
& & fires to generate peristalsis\\[5pt]
B10 & Medial/posterior I3 & Fixed time delay after B31/32\\[5pt]
\multirow{2}{*}{B44/B48} & \multirow{2}{*}{Odontophore Opener} & On for protraction in biting/swallowing\\
& & On for retraction in rejection\\[5pt]
B43/B45 & I1 & On for retraction in all behaviors\\
\bottomrule
\end{tabularx}
\label{tab:neurons}
\end{center}
\vspace{-25pt}
\end{table}

Additional neurons and connections were added (Table~\ref{tab:neurons}) to the existing model to activate muscles that were included in the robot but not in the previous biomechanical model~\cite{Webster-wood2020ControlCalifornica}. Motor neurons were needed for the I1 muscles, the odontophore opener, and the individual segments of I3. Note that these additional interneuron connections are based on phenomenological connections and may not reflect the true neural connections. For I1, motor neurons B43 and B45 were selected~\cite{Church1991ExpressionAplysia}, and an excitatory connection with B64 was hypothesized to synchronize firing with retraction. The odontophore opener had behavior-dependent excitation from B48 (ingestive, firing during retraction) and B44 (egestive, firing during protraction)~\cite{Cropper2019TheAplysia}. As the opener is a single muscle, these neurons were lumped together and excited by B64 to mark retraction and by CBI-3 to indicate ingestion or egestion. The I3 required a decohered B3/B6/B9 motor pool to produce peristalsis, which was complicated by reports of different firing patterns in intracellular recordings~\cite{Church1994ActivityAplysia, gill_rapid_2020}. For generating peristalsis in SLUGBOT, we introduced fixed time delays between the sequential firings for three functional retractor motor units that may not reflect true firing patterns of B3, B6, and B9 \textit{in vivo}.
Finally, the context-dependent role of I3 as a protractor~\cite{Sutton2004NeuralAplysia} was incorporated into the design, so additional innervation to the posterior of I3 was performed by B10~\cite{Church1991ExpressionAplysia} near the end of protraction~\cite{Morton1993TheAplysia}, activated by a fixed time delay after the firing of B31/B32. The activation of each I3 ring was set as a weighted sum of three Retractor Units (RU1/RU2/RU3), B10, and B38 (responsible for pinching in the anterior I3 \cite{McManus2014DifferentialFunction}) based on the regional innervation~\cite{Church1991ExpressionAplysia}. The weights were experimentally tuned to produce peristalsis. 

During experimentation, the neural signals were calculated in real time on one Teensy 3.6, and then passed to the pressure microcontroller via UART. The neural controller collected the proprioceptive feedback from the integrated sensors, calculated the Boolean neural signals, and logged the neural and sensors signals to a PC. Different SLUGBOT behaviors could be set by changing the states of sensory feedback neurons (related to mechanical stimulus at the lips, chemical stimulus at the lips, and mechanical stimulus at the grasper) as well as the general arousal of the grasper via  push buttons. During swallowing, tape strips or plastic tubing were placed in the grasper's radula to represent seaweed.

\vspace{-10 pt}
\subsubsection{Muscle Pressure Controller}

The Boolean neuron signals were received from the neural microcontroller and integrated by a muscle pressure microcontroller (Teensy 3.6). These activations were scaled and used as pressure set points. For each actuator, pressurization was controlled through bang-bang actuation of two electromechanical valves (KOGANEI GA101HE1) with feedback from a 30~psig pressure sensor (ELVH-030G-HAND-C-PSA4). One valve acted as an inlet leading to the pressure supply line (15~psig) and another valve acted as relief leading to ambient pressure. Pressure information from each of the ten actuators was collected serially, with actuation done in parallel through three daisy-chained shift-register low-side switches (TPIC6595). A 0.4-psig tolerance band was added around the pressure setpoint to further improve stability.

\vspace{-10 pt}
\section{Results}
\vspace{-10 pt}

\begin{wrapfigure}[30]{R}{0.65\textwidth}
\vspace{-35pt}
    \includegraphics[width=0.65\textwidth]{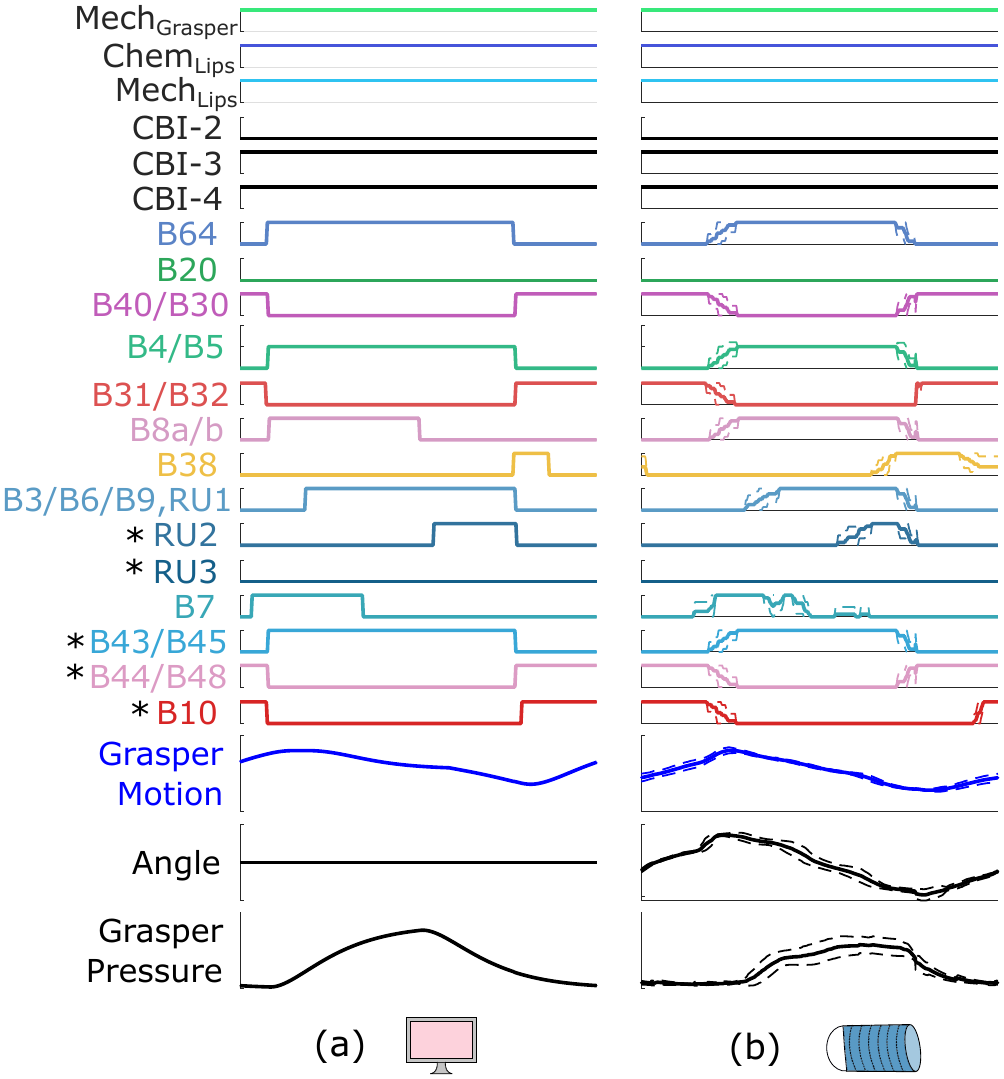}
    \setlength{\abovecaptionskip}{-15pt}

    \caption{SLUGBOT qualitatively replicates the \textit{in silico}~\cite{Webster-wood2020ControlCalifornica} neural signals, kinematics, and grasper pressure. SLUGBOT data is normalized to cycle length and averaged over multiple cycles, with dashed lines indicating standard deviation. a) Boolean model, swallowing; b) SLUGBOT, swallowing ($n=8$). Asterisks indicate neural signals modified from the hybrid Boolean model. Retractor Unit 1 follows the same activation logic as B3/B6/B9. $\text{Mech}_\text{i}$ indicates mechanical stimulus at location $i$ (either the grasper or lips, and $\text{Chem}_\text{Lips}$ indicates chemical stimulus at the lips.}
    \label{fig:results}
\vspace{-15pt}
\end{wrapfigure}

\begin{wrapfigure}[31]{R}{0.59\textwidth}
\vspace{-45pt}
    \centering
    \includegraphics[width=0.55\textwidth]{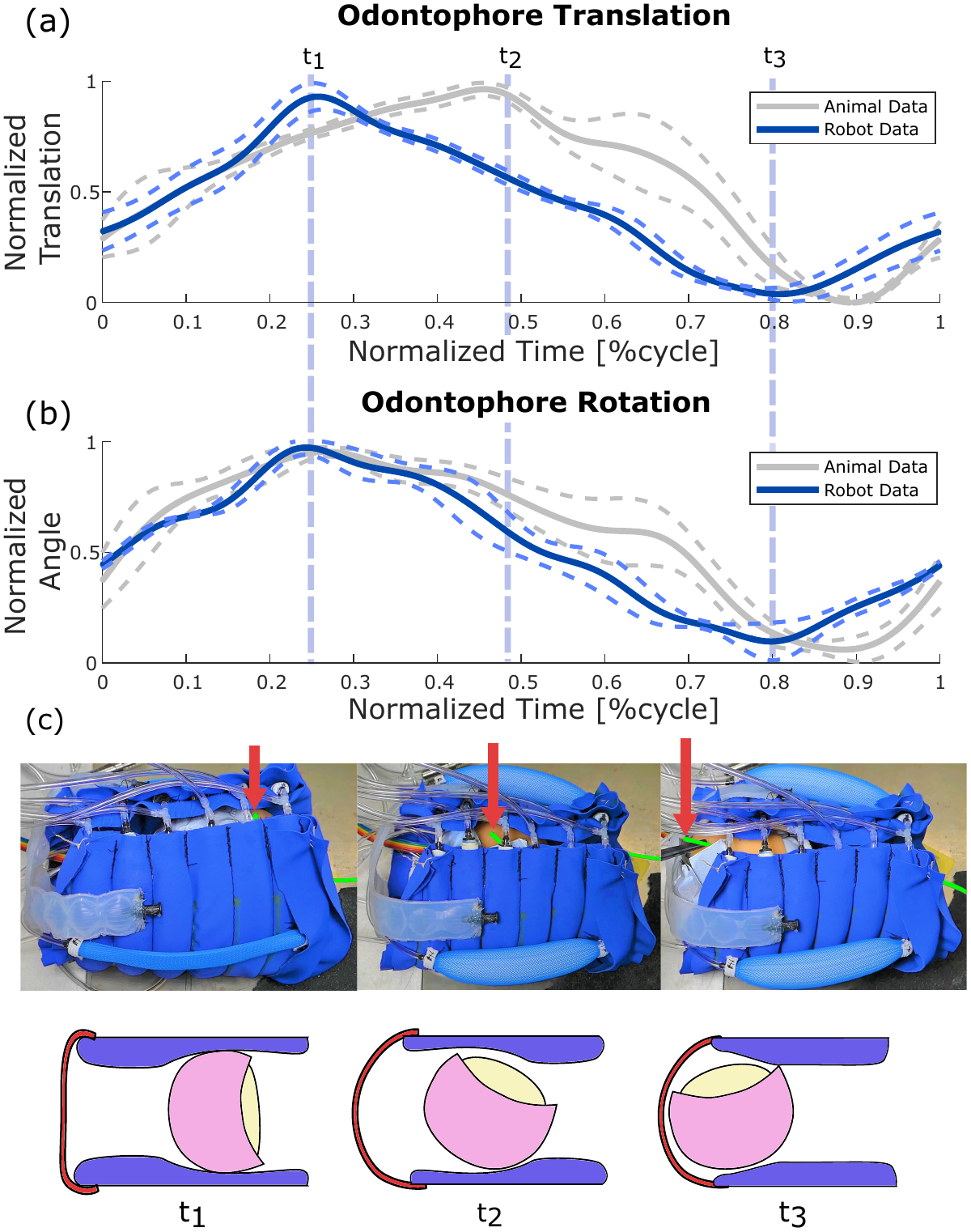}
    \vspace{-10pt}
    \caption{
		Kinematic comparison with MRI data from behaving animals. (a) Odontophore translation. (b) Odontophore rotation. (c) Frames from swallowing video with corresponding cartoon representation showing the retraction phase and the "ingestion" of a tube. The time-normalized swallowing kinematics are compared between the robot and the behaving animal data taken from MRI images, presented in~\cite{Neustadter2002AImages}. Three time points in the retraction phase (peak protraction, mid retraction, and peak retraction) are indicated in the data, and corresponding images and cartoon are shown. A small tube (highlighted in green) was fed to the robot and successfully transferred to the rear of the robot. The tip of the tube is indicated with an arrow. 
    } \label{fig:consistency}
\vspace{-20pt}
\end{wrapfigure}

SLUGBOT qualitatively demonstrated swallowing behavior when presented with the corresponding stimuli (Figures~\ref{fig:results} and ~\ref{fig:consistency}). The data are time shifted to align peak retraction with peak retraction reported in \cite{Neustadter2002AImages}. Multiple cycles ($n=8$) were time normalized and averaged together, using the activation of B31/B32 to determine the normalization period. Due to this averaging, some robot neural signals show values within the range of 0 to 1, rather than Boolean values of 0 and 1. The additional neurons from the robot controller were implemented in the original Boolean model framework for comparison but do not affect the \textit{in silico} biomechanics. For most neural signals, activation and deactivation triggered at similar points in the cycle, with a range of 13.3\% of cycle length for swallowing. In both models, a third retractor unit, RU3, was implemented but did not fire because the fixed time delay after RU1 fired was longer than the retraction time.

In comparison to the \textit{in vivo} kinematics of \textit{Aplysia's} odontophore measured by Neustadter et al. \cite{Neustadter2002AImages}, SLUGBOT reproduces the main characterisitics of the odontophore's translation and rotation during swallowing. For both \textit{in vivo} and \textit{in roboto} swallows, the forward translation of the odontophore during protraction is accompanied by simultaneous rotation. The radula rotate approximately 90 degrees during protraction, with the radula facing the dorsal surface at the start of the cycle and then facing outwards from the lumen at peak protraction~\cite{neustadter_kinematics_2007}. The odontophore then retracts by translating towards the posterior to bring the grasped food into the lumen and towards the esophagus.  This is accompanied by an approximately 90 degree rotation of the odontophore such that the radula are facing the dorsal surface of the buccal mass, as with the start of protraction. 

\vspace{-10 pt}
\section{Discussion}
\vspace{-5 pt}
\subsection{Morphological differences between SLUGBOT and \textit{Aplysia}}
\vspace{-5 pt}
SLUGBOT qualitatively replicated key features of the translational and rotational kinematics of the \textit{Aplysia's} odontophore measured during \textit{in vivo} swallowing experiments.  Once the parameters for the Boolean controller were tuned, the range and timing of the odontophore's translations and rotations were repeatable during the cyclic protraction and retraction during swallowing. 

However, there exists a number of key morphological features present in the animal which are essential for successful feeding behavior but were not captured in SLUGBOT.  In the animal, the anterior portion of the I3 musculature near the jaws tightens and clamps together, holding food in place during swallowing \cite{McManus2014DifferentialFunction}. The toroidal McKibben actuators that constituted SLUGBOT's I3 musculature were not capable of sufficient radial expansion to clamp onto the narrow plastic tubing representing food.  As a consequence, the plastic tubing needed to be held in place by methods external to the robot, which is not needed \textit{in vivo}. 

\textit{In vivo}, the odontophore is capable of a total range of posterior rotation beyond 90$^{\circ}$ during swallowing \cite{Neustadter2002AImages, Novakovic2006MechanicalCalifornica}. This additional rotation was utilized by Mangan et al. to translocate an object through their \textit{Aplysia}-inspired grasper~\cite{mangan2005biologically}. SLUGBOT's inability to rotate beyond 90$^\circ$ may have contributed to inconsistent swallowing behavior because the radular surface was not tilted towards the esophagus as in \textit{Aplysia}. Plastic tubing released at the end of retraction tended to fall directly back onto the radular surface instead of being deposited posterior to the robot. \textit{In vivo}, both active and passive hinge forces play an important role in assisting retraction~\cite{Sutton2004PassiveSwallowing}. Modification of the hinge's mechanical properties and adding active control may enable more biomimetic motion. 

The deformation of the soft structures and musculature of the buccal mass plays an important role \textit{in vivo} that can improve SLUGBOT's biomimicry and feeding performance in the future.  For instance, the movement of the radular stalk into the odontophore both increases posterior rotation during the loss of the $\Gamma$ shape and also changes the effective shape of the odontophore \cite{Novakovic2006MechanicalCalifornica}.  The shape of the odontophore during retraction increases the mechanical advantage and effective force from the I1/I3 complex, enhancing retraction~\cite{Novakovic2006MechanicalCalifornica}. The I2 muscle likewise benefits from mechanical reconfiguration to enhance protraction during rejection and biting~\cite{Novakovic2006MechanicalCalifornica}. SLUGBOT is not currently capable of taking advantage of such mechanical reconfiguration because (1) the odontophore is a rigid sphere and (2) there is no analogue to the radular stalk. Using softer materials and incorporating a radular stalk-like mechanism can help address this gap. The context-dependent enhancement of the effective force on the odontophore could reduce the impacts of friction and poor mechanical advantage during movement.

\vspace{-8 pt}
\subsection{Neuronal Controller}
\vspace{-5 pt}
By implementing a biomimetic neuronal controller using Boolean logic to control SLUGBOT, we were able to demonstrate swallowing. Qualitatively, the behavior resembled that of the previously developed \emph{in silico} Boolean model~\cite{Webster-wood2020ControlCalifornica} when comparing the neural signals, grasper motion, and grasping force. Repeatability between cycles and comparable motion profiles suggest that bioinspired control of \textit{Aplysia}-inspired robotic structures is possible even with simplified neural mechanics and a smaller number of controlling neurons than in the animal. Both the \textit{in silico} and \textit{in roboto} models are qualitative models of the grasper motion and are not fit to animal data. As the appropriate \textit{in vivo} data is collected, the robot can be further tuned to match the observed behaviors, such as with the different activation patterns of B3, B6 and B9 neurons recorded \textit{in vivo} \cite{Church1994ActivityAplysia, gill_rapid_2020}. 

\vspace{-5 pt}
\section{CONCLUSION}
\vspace{-5 pt}
SLUGBOT - a soft robotic grasper inspired by the buccal mass of  \emph{Aplysia californica} -- demonstrates robotic control via Boolean logic to replicate swallowing behavior exhibited by the model animal. The robot's grasping motion and force profiles qualitatively resemble that of the previously developed \emph{in silico} Boolean model.  Future iterations of SLUGBOT can improve its feeding performance by incorporating additional biomimetic elements that are not present in the current version.  Of particular importance are the anterior pinch of the I3 musculature to hold food in place, mechanical reconfiguration of the shape of the odontophore and better matching of the hinge and I2 properties to those measured \textit{in vivo}. Further developments could also improve SLUGBOT's viability as a platform for neuromechanical research by enabling behavorial switching and integration with more complex neural models. This brings \textit{Aplysia}-mimetic control \emph{in roboto} one step closer to multifunctional neural control schema \emph{in vivo} and \emph{in silico}.

\vspace{-5 pt}
\section*{ACKNOWLEDGMENT}
\vspace{-7 pt}
We thank Jesse Grupper (Harvard University) and Al Turney (KOGANEI International America, Inc.) for help in developing the pneumatic controller.

\vspace{-7 pt}

\def\url#1{}
\bibliographystyle{splncs04}
\bibliography{ref}

\end{document}